\documentclass[11pt,a4paper]{article}
\usepackage[hyperref]{emnlp-ijcnlp-2019}
\usepackage{times}
\usepackage{latexsym}
\usepackage{amsmath}
\usepackage{booktabs}
\usepackage{color}
\usepackage{graphicx}
\usepackage{svg}
\usepackage{textcomp}
\usepackage{times}
\usepackage{url}
\usepackage{caption}
\usepackage{siunitx}
\usepackage{gensymb}

\aclfinalcopy 

\newcommand*{\round}[1]{\num[round-mode=places,round-precision=1]{#1}}

\title{Using Local Knowledge Graph Construction to \\ 
Scale Seq2Seq Models to Multi-Document Inputs}

\author{Angela Fan \\
  FAIR / LORIA \\
  \And
  Claire Gardent \\
  CNRS / LORIA \\
  \And
  Chlo\'e Braud \\
  CNRS / LORIA \\
  \And
  Antoine Bordes \\
  FAIR 
  \AND 
  {\tt {angelafan,abordes}@fb.com} \\
  {\tt {claire.gardent,chloe.braud}@loria.fr}}

\date{}

\begin{document}
\maketitle 
\begin{abstract}
Query-based open-domain NLP tasks require information synthesis from long and diverse web results. Current approaches extractively select portions of web text as input to Sequence-to-Sequence models using methods such as TF-IDF ranking. We propose constructing a local graph structured knowledge base for each query, which compresses the web search information and reduces redundancy. We show that by linearizing the graph into a structured input sequence, models can encode the graph representations within a standard Sequence-to-Sequence setting. For two generative tasks with very long text input, long-form question answering and multi-document summarization, feeding graph representations as input can achieve better performance than using retrieved text portions.
\end{abstract}

\section{Introduction}

Effective information synthesis is at the core of many Natural Language Processing applications, such as open-domain question answering and multi-document summarization. In such tasks, a fundamental challenge is the ability to distill relevant knowledge from hundreds of thousands of tokens of noisy and redundant input such as webpages. Current approaches predominantly conduct TF-IDF-based information extraction to identify key portions of the information, and then provide this as sequential input to a Sequence-to-Sequence (Seq2Seq) model. The sub-selected portions are limited to a few thousand words, as models often struggle to encode much longer sequences.

In this work, we propose a mechanism to re-structure free text into local knowledge graphs that are then linearized into sequences, creating a canonical form in which information is presented to models. By constructing a graph intermediary, redundant information can be merged or discarded, producing substantially compressed input --- small enough to be fully encoded by Seq2Seq models. Such a method can be seen as merging previous work on symbolic knowledge bases for information extraction with newer approaches using deep neural networks to encode knowledge. 

\begin{figure}[t!]
    \centering
    \includegraphics[width=\linewidth]{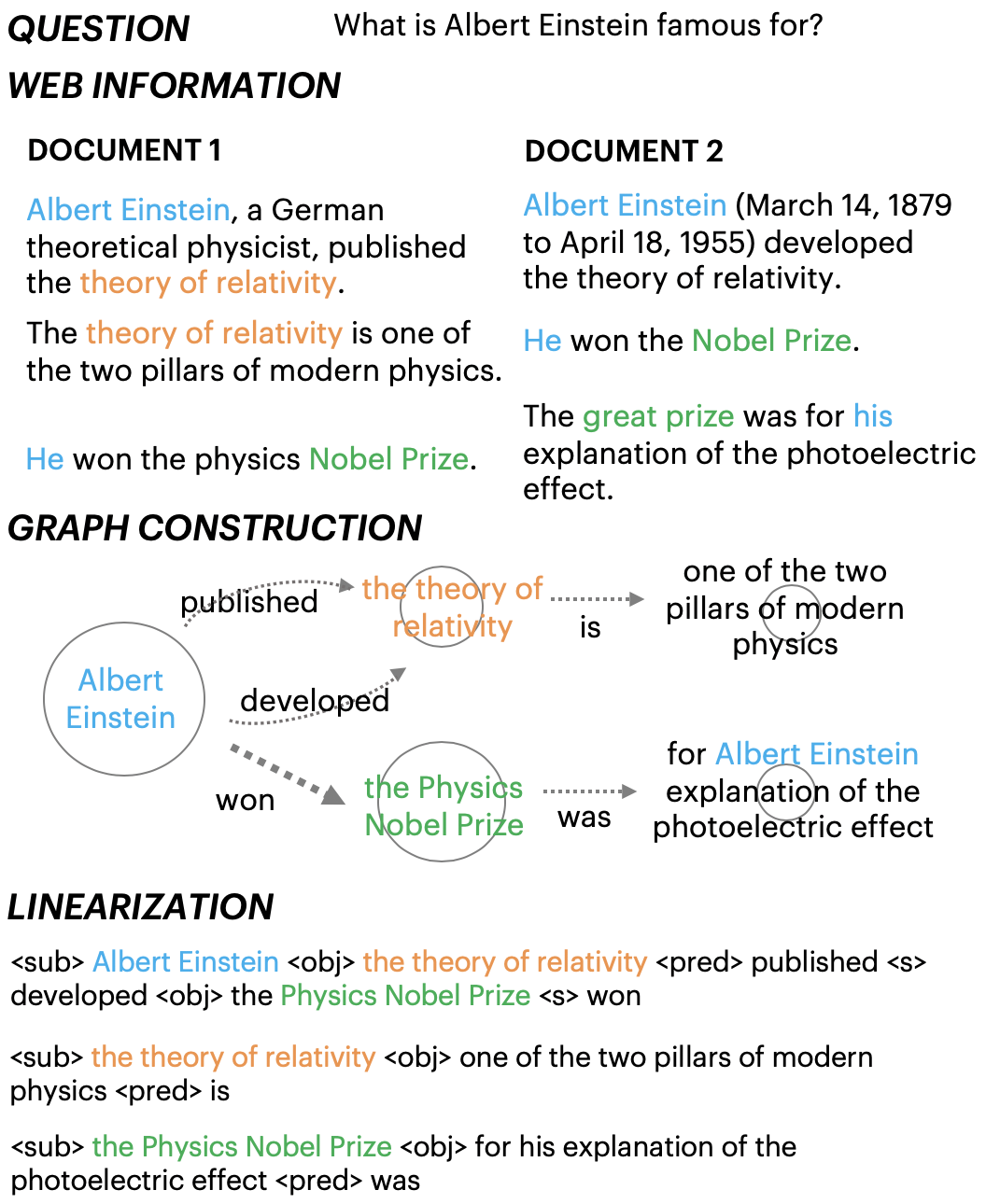}
    \caption{\textbf{Multi-Document Input to Linearized Graph} Multi-document input resulting from web search queries are converted to a graph structured knowledge base using coreference resolution and information extraction, then linearized into a sequence for Seq2Seq models. Color indicates coreference resolution. Node weight is indicated by circle radius and edge weight by line thickness.}
    \label{fig:example}
\end{figure}

Our approach, shown in Figure~\ref{fig:example}, takes a query and its corresponding multi-document web search results and builds for each query a specific local knowledge graph. We present several modeling contributions to effectively encode the entire graph as a sequence and attend to the most relevant portions within this linearization. We demonstrate the effectiveness of this approach on two large-scale generative tasks with both long and noisy multi-document web input and paragraph length output: long-form question answering on the ELI5 dataset \cite{fan2019explain} and Wikipedia lead paragraph generation as a multi-document summarization problem \cite{liu2018generating}. 

\section{Related Work} 

Interest in generative sequence modeling has intensified due to recent improvements \cite{peters2018deep,devlin2018bert,radford2019language}, making the challenge of information synthesis more relevant. In contrast to extractive tasks which only require models to identify spans and can do so effectively on long documents by looking at the paragraphs independently, generative sequence models must combine multiple pieces of evidence from long and noisy multi-document input to generate correct and convincing responses.

\subsection{Multi-Document Input}

Previous work in multi-document summarization \cite{barzilay1999information} applies various techniques to handle long input, including clustering to find similar information \cite{honarpisheh2008multi}, extractive methods to select relevant sentences \cite{daume2002noisy,gillick2009scalable,berg2011jointly,di2014hybrid,bing2015abstractive,cao2017improving} including maximal marginal relevance \cite{fabbri2019multi}, and incorporating queries \cite{baumel2018query} and graphs \cite{ganesan2010opinosis,yasunaga2017graph}. However, there are few large scale multi-document summarization datasets and many approaches have focused on extractive selection or hybrid extractive-abstractive models. In this work, we use graph construction to re-structure multi-document input for abstractive generation. 

Advancements in question answering have examined performance on datasets with multi-document input, such as TriviaQA \cite{joshi2017triviaqa}. Various approaches have been proposed, including leveraging TF-IDF and bigram hashing with an RNN to find relevant information \cite{chen2017reading}, models that score individual paragraphs for sub-selection \cite{clark2017simple}, and nearest neighbor search with paragraph re-ranking \cite{das2018multi}. However, these approaches have been applied to extractive question answering tasks that require span identification, rather than abstractive text generation in an information synthesis setting. 

\subsection{Using Knowledge Bases}

Previous work has explored various ways of representing information in knowledge bases \cite{bordes2011learning} and improving these representations \cite{chen2013learning}. Knowledge bases have been leveraged to improve performance on various tasks, from coreference resolution \cite{ng2002improving} and question answering \cite{zheng2003question,bao2014knowledge,cui2017kbqa,sun2018open} to signal processing \cite{buckner2002knowledge}. Various works convert text into Abstract Meaning Representations \cite{liu2018toward} for domains such as news \cite{vossen2015storylines,rospocher2016building} and link nodes to large knowledge bases such as DBPedia \cite{auer2007dbpedia}.  \citet{wities2017consolidated} combine open information extraction with coreference and lexical inference to build knowledge representations. They apply this to tweets and analyze the accuracy on various aspects of graph construction. \citet{das2018building} construct graphs from procedural text to track entity position to answer when and if entities are created, destroyed, or moved. In contrast, we build graphs from substantially longer multi-document input and use them for multi-sentence text generation.

Recently, many have explored neural architectures that can encode graph structured input \cite{bruna2013spectral,kipf2016semi,beck2018graph,zhou2018graph,xu2018graph2seq,lai2019lattice}. These models often represent graphs as adjacency matrices to generalize architectures such as convolutional networks to graph inputs. Rather than encoding a static knowledge graph or leveraging external knowledge graphs, we build a local graph for each query and model these using standard Seq2Seq models. We leave the incorporation of graph networks for future work. 

\section{Graph Construction}

We describe how symbolic graph representations of knowledge can be constructed from text. Our approach assumes a multi-document input (such as web pages) resulting from the execution of a query. The graph construction process (1) compresses the web search input to a significantly smaller size, allowing  models to encode the entirety of the compression, and (2) reduces redundancy through merge operations, allowing relevant information to be more easily identified. 

\begin{figure}[t!]
    \centering
    \includegraphics[width=\linewidth]{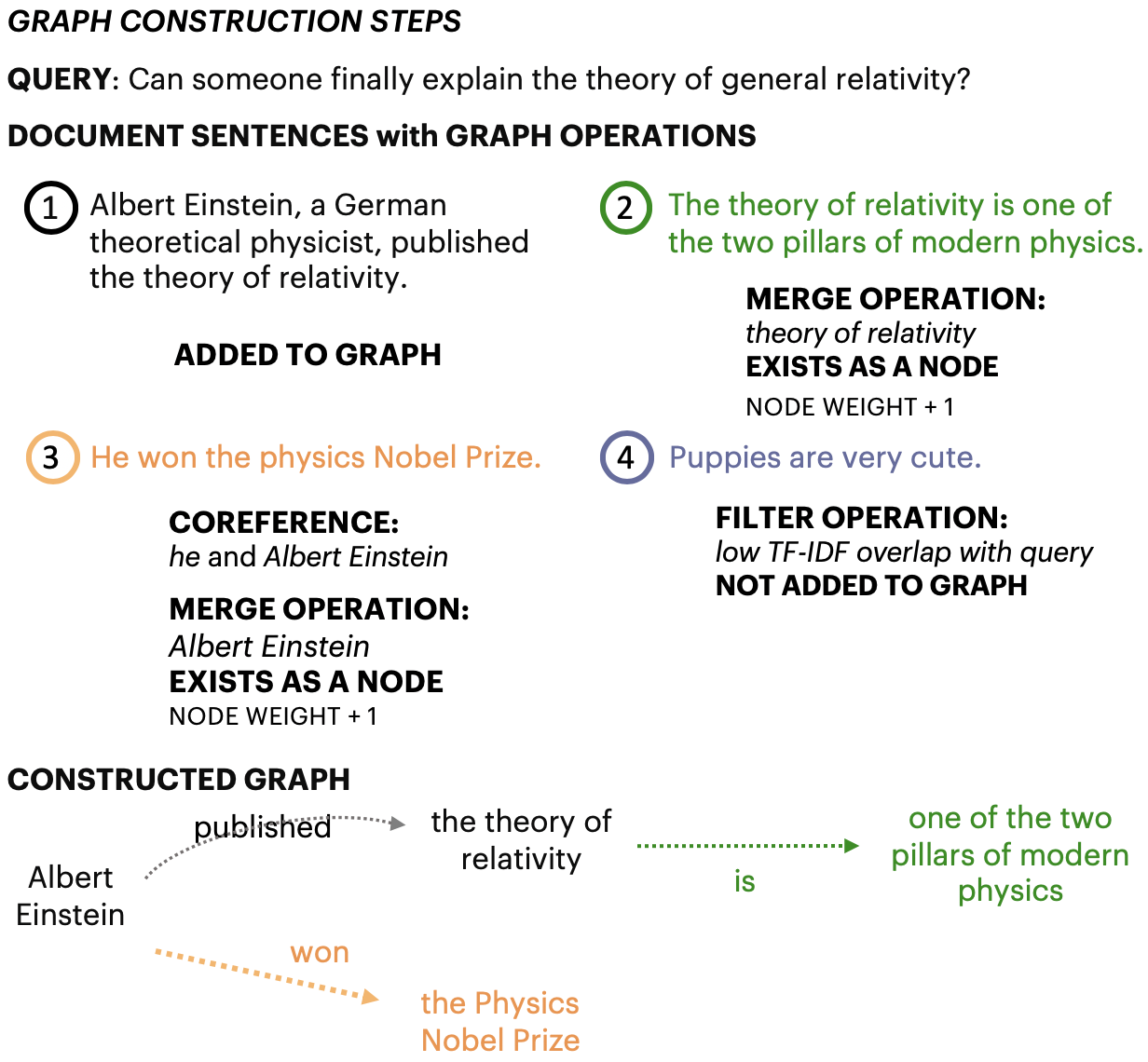}
    \caption{Steps of \textbf{Graph Construction}. Color relates the document sentence used to produce the graph output.}
    \label{fig:construction}
\end{figure}

\paragraph{Text to Triples to Graph} Graph construction proceeds in several steps outlined in Figure~\ref{fig:construction}. We apply \textit{Coreference Resolution} \cite{clark2016deep,clark2016improving}\footnote{We use the implementation available here: \url{https://github.com/huggingface/neuralcoref}} and \textit{Open Information Extraction} \cite{stanovsky2018supervised}\footnote{We use the implementation available here: \url{https://github.com/gabrielStanovsky/supervised-oie}} to convert sentences into a \textit{Triple} of the form (subject, predicate, object). The sentence \textit{Albert Einstein, a German theoretical physicist, won the Nobel Prize} would become (\textit{Albert Einstein, won, the Nobel Prize}). 

A graph is constructed using the triples by representing subjects and objects as nodes connected by predicates as directed edges. For example, the triple would become \textit{Albert Einstein} $\xrightarrow[\text{won}]{}$ \textit{the Nobel Prize}. Nodes and edges have a \textit{name} property that is the text they represent. They also have a \textit{weight} property that denotes the number of times that node or edge has appeared. For example, in Figure~\ref{fig:example}, the node with name \textit{Albert Einstein} has weight 4 and edge with name \textit{won} has weight 2. 

\begin{figure*}[t!]
    \centering
    \includegraphics[width=\linewidth]{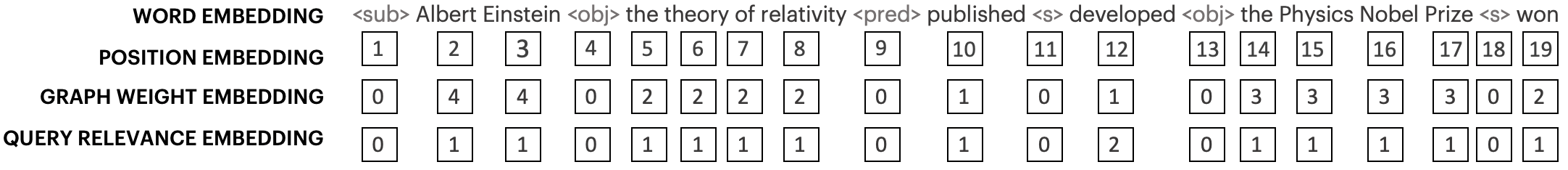}
    \caption{\textbf{Graph Attribute Embeddings}. In addition to word and position embeddings, models receive a Graph Weight embedding to encode node and edge weight and a Query Relevance embedding that encodes search result rank.}
    \label{fig:embed_example}
\end{figure*}

\paragraph{Merging Nodes and Edges} When subsequent triples are added to the graph, they are merged with the existing graph if they already exist to reduce information replication. To merge nodes, the TF-IDF overlap of the new node's name is calculated with the existing graph node names, and the new node is merged into an existing node if the TF-IDF is higher than some threshold (see Figure~\ref{fig:construction}, steps 2 and 3 for example merge operations). Edges are merged similarly with existing edges between the same two nodes. Such merge operations allow strings such as \textit{the Nobel Prize} and \textit{Nobel Prize} to be represented as one node rather than separately. Similarly,  coreference resolution aids in merging --- by identifying that \textit{Albert Einstein} and \textit{He} refer to the same entity and thus merging them, the construction of the graph reduces redundancy. 
The size of the graph can be modified by controlling which triples are added using TF-IDF overlap (see Figure~\ref{fig:construction}, step 4). TF-IDF overlap of the triple with the question can be used to determine if the triple contains relevant information. This improves robustness to noisy web search input and helps filter entirely irrelevant portions, such as scraped HTML tags.

\section{Modeling Graphs as Sequences}

\begin{figure}[t!]
    \centering
    \includegraphics[width=\linewidth]{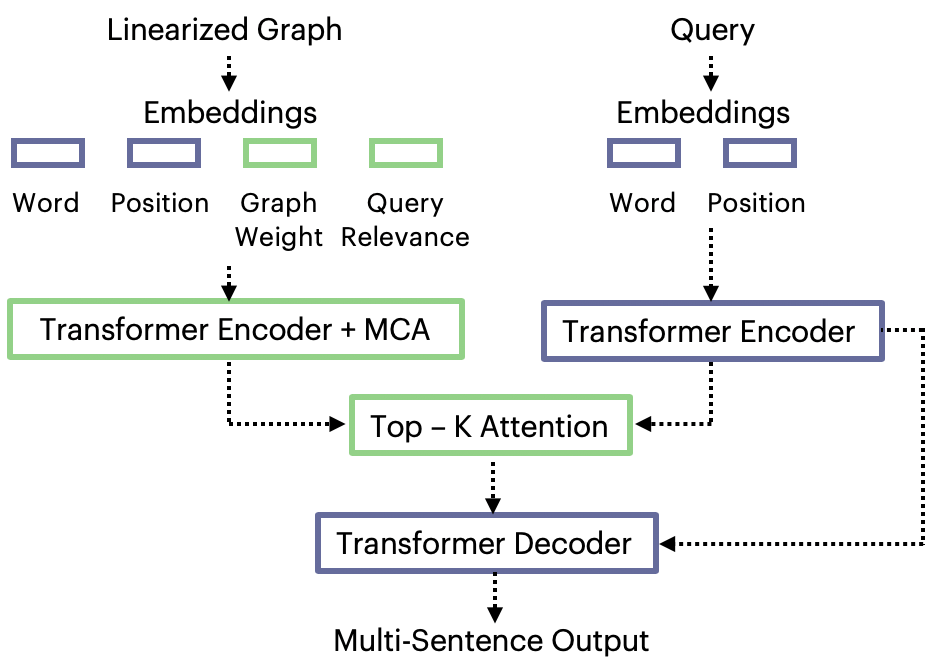}
    \caption{\textbf{Model Architecture}. Gray indicates standard Transformer elements, Green indicates modification.}
    \label{fig:model_architecture}
\end{figure}

Current models for text generation often use Seq2Seq architectures such as the Transformer \cite{vaswani2018attention}. These models are designed to encode sequences rather than graphs. We describe now how to convert a graph into a structured input sequence. Our complete model will take as input a linearized graph by encoding graph attributes such as node and edge weight as embeddings. We add hierarchical and memory-compressed attention mechanisms to scale Seq2Seq models to encode the full graph and attend to the most relevant information within it (Figure~\ref{fig:model_architecture}), and finally we integrate the benefits of language modeling using multi-task training. 

\subsection{Graph to Sequence} 

\paragraph{Linearization} To represent the graph as a sequence for Seq2Seq, it is linearized into a structured standard form of subject node, object node, and predicate edge, separated by indicator tokens, as shown in Figure~\ref{fig:example}. For example, the linearization \textit{$<$sub$>$ Albert Einstein $<$obj$>$ the Nobel Prize $<$pred$>$ won} would be created. The linearization is accomplished through graph traversal in a breadth-first manner following the directed edges formed by predicates and starting from the node with the largest weight as the root. For two nodes that are connected by multiple predicates, the predicates are concatenated (shown in Figure~\ref{fig:example}), so a linearization such as \textit{$<$pred$>$ won $<$s$>$ received} would indicate that Albert Einstein both won and received the Nobel Prize. 

\paragraph{Encoding Graph Information} Transformer Seq2Seq models have two embeddings: a word embedding and a position embedding. Simply linearizing the graph, however, loses attribute information such as node and edge weight. Instead, we encode these attributes as embeddings in addition to the word and position embeddings. 

To represent \textit{Graph Weight} (GW), node and edge weight is provided as an embedding for each token. The node weight and edge weight are equivalent to the number of merge operations + 1. For example, if \textit{Albert Einstein} occurred 4 times in the text, the GW embedding for the tokens \textit{Albert} and \textit{Einstein} would be 4, as shown in Figure~\ref{fig:embed_example}. 

We encode a \textit{Query Relevance} (QR) embedding to represent the relevance of the web search to the query as ranked by the information retrieval system (e.g. search engine). Information from the top web search results is likely more relevant than information from the last web search results, so providing this embedding allows the model to distinguish between these different information sources. In Figure~\ref{fig:embed_example}, tokens representing sentences from the first document have QR embedding 1, and tokens from the second document have value 2. 

Models now have access to several different types of embeddings, but all embedding information contributes equally as there is no mechanism to distinguish between them. We introduce a mechanism for models to scale the graph embeddings. We denote the embedding for position $t$ as  $e_t$, such that $e_t^{\textrm{word}}$ is the word embedding.

For the GW embedding, models learn a gating function $g$ based on the word and GW embeddings. Such a mechanism provides capacity for the model to decide when the additional embeddings are useful based on the words in the input. The gate is calculated by applying an MLP $W$ to the concatenation of the word and GW embeddings. The learned gate is then applied to GW embeddings to create the output $h$:
\begin{align}
g_t^{\textrm{GW}} &= W [e_t^{\textrm{GW}}; e_t^{\textrm{word}}]\nonumber\\
h_t^{\textrm{GW}} &= g_t^{\textrm{GW}}\circ e_t^{\textrm{GW}}\nonumber
\end{align}

Models learn a gating mechanism for the QR embedding in a similar manner. The full embedding the model receives is as follows:
\begin{align}
e_t^{\textrm{word}} + e_t^{\textrm{pos}} + [h_t^{\textrm{GW}}; h_t^{\textrm{QR}}] \nonumber
\end{align}

\subsection{Hierarchical Attention} One challenge in modeling long sequences is that attention mechanisms struggle to make sharp selections when softmax-ing over long sequences \cite{fan2018hierarchical}. When attention is blurry, there lacks a strong distinction between noise and relevant information.

We assume that graphs are constructed from query-based web search input and thus leverage this query to learn a subselection operation using \textit{hierarchical top-$k$ attention}, depicted in Figure~\ref{fig:model_architecture}. The query is encoded with a Transformer encoder and the linearized graph with another Transformer encoder. We model the interaction between the query and the input sequence (e.g. web search results or linearized graph) by computing an attention distribution between the question and the input, then selecting the top $k$ positions with the most attention weight. Such a mechanism can be thought of as building a query-dependent representation of the most relevant knowledge, which is commonly done in question answering architectures like BiDAF \cite{seo2017bidirectional}. The top $k$ operation limits the number of tokens, making the attention mechanism sharper. 

\subsection{Scaling to Encode the Graph} 

Recent progress has improved the ability of language models to process longer sequences \cite{sukhbaatar2019adaptive,dai2019transformer}, but models remain limited in their capacity to encode long documents. The multi-document results of query-based web search have hundreds of thousands of tokens, beyond the limit of current Seq2Seq models to handle. For example, the ELI5 dataset provides an average of 200K tokens of web search input. However, by compressing the web search results into a knowledge graph, we significantly reduce the number of tokens by an order of magnitude and make it possible for a model to access the entirety of the search information.

To represent the full graph, models must scale to encode around 10K input tokens. The attention mechanism in Transformer architectures becomes  computationally expensive for sequences of this length. Instead, we experiment with the \textit{Memory-Compressed Attention} (MCA) mechanism proposed for language models in \cite{liu2018generating}\footnote{In \cite{liu2018generating}, the mechanism is termed \textit{DMCA} as it is applied on the decoder side} and apply it to the encoder side of Seq2Seq models. At each self-attention layer, MCA alternates between (1) local attention, computed between smaller chunks of tokens and (2) strided convolutions to reduce the number of keys and values used in attention computation. By adding the MCA mechanism to the encoder (\textit{E-MCA}), we are able to encode the complete linearized graph.

\subsection{Multi-tasking with KB Completion} \citet{fan2019explain} used multi-task training on language modeling and various Seq2Seq tasks to incorporate the benefits of language modeling in Seq2Seq models. We extend this by training additionally on \textit{knowledge graph completion}. Models receive at training time sequences of a linearized graph with nodes, edges, or both selectively masked and must predict the missing content words. For example, models might receive as input \textit{$<$sub$>$ Albert Einstein $<$obj$>$ [mask] $<$pred$>$ won} and need to predict \textit{the Nobel Prize}. This can be seen as both a multi-word extension of the masked language model training proposed in \cite{devlin2018bert} and as learning the task of knowledge base completion \cite{lacroix2018canonical,bordes2011learning}. At training time, the tasks are distinguished using an indicator token in the input. 

\section{Experimental Setup} 

We evaluate our approach on two datasets with multi-document web input and multi-sentence abstractive output. We use Seq2Seq models that leverage a  query concatenated with web search results that have been processed into a supporting document --- e.g. TF-IDF subselection, linearized graph, etc. --- to generate long output.  

\subsection{Datasets and Evaluation} 

\paragraph{ELI5} First, we experiment with the \textit{Explain Like I'm Five} (ELI5) \cite{fan2019explain} question answering dataset of 270K complex questions paired with multi-sentence, explanatory answers (130 words on average). To facilitate question answering, the dataset provides the top 100 web search hits from querying the question, which results in 200K words on average. See Appendix for examples. 

Previous work \cite{fan2019explain} used TF-IDF to find sentences in the web search that have the largest overlap with the question and created a \textit{TF-IDF extraction} of about 850 words as input for Seq2Seq models. Instead, we construct a local knowledge graph for each question from the 100 web search hits. Following the average length of the TF-IDF support document constructed in \cite{fan2019explain}, we experiment with modeling the first $N = 850$ tokens of the linearized graph, then scale to encode the entire graph using E-MCA. 

\paragraph{WikiSum} Second, we experiment on the \textit{WikiSum CommonCrawl} \cite{liu2018generating} summarization dataset\footnote{\url{https://github.com/tensorflow/tensor2tensor/tree/master/tensor2tensor/data_generators/wikisum}} with 1.5 million examples. This task formulates Wikipedia lead paragraph generation as a multi-document summarization problem, where the paragraph must be generated using the cited article references and other queried content from web search. The query used is the title of the Wikipedia article. See Appendix for examples.

Previous work \cite{liu2018generating} applied \textit{TF-IDF Ranking} to order the paragraphs of web search given a query. Models receive the re-ordered paragraphs ranked by TF-IDF as input. \citet{liu2018generating} model the first $N=500$ words of this re-ranking and then $N=11,000$ using MCA. We construct the knowledge graph for each Wikipedia article from the first 200K words of the ranked web search results\footnote{Average length of provided web input is around 50K words, and maximum length is around 900K words}, and experiment with encoding $500$ and $11,000$ tokens.

\paragraph{Evaluation} Both tasks evaluate the multi-sentence generation output against the gold output using F1 ROUGE. On WikiSum, we report only ROUGE-L following \cite{liu2018generating}. 

We conduct a comparative human evaluation on the ELI5 dataset. We use crowdworkers to examine the responses of two models on 300 different questions from the test set. For each question, 3 evaluators are shown two answers and asked to choose the one they prefer. To reduce variance, answers are standardized at 150 words each. 

\subsection{Training and Generation}

To reduce the vocabulary size of varied web document content, we apply byte-pair encoding \cite{sennrich2016neural} to generate 40K codes for each dataset. We implement our models in fairseq-py \cite{ott2019fairseq} using the Transformer Big architecture and training schedule described in \cite{vaswani2018attention}. Detailed parameters are listed in the Appendix. For generation, we use beam search with beam size 5 and tune a minimum and maximum length on the validation set.

\subsection{Baselines}

We compare our results to the Transformer sequence models presented in \cite{fan2019explain} for ELI5 and \cite{liu2018generating} for WikiSum. 

We evaluate three additional baseline models: 
\begin{itemize}
    \setlength\itemsep{-0.2em}
    \item \textit{Sentence Selection with Maximal Marginal Relevance:} \cite{fan2019explain} used TF-IDF to identify relevant sentences in the web documents to form a support document of around 850 words. However, recent work \cite{fabbri2019multi} has shown that using maximal marginal relevance is an effective strategy for selecting relevant information while reducing redundancy. We explore using MMR to select sentences from the web text to concatenate to form a support document.
    \item \textit{Seq2Seq Multi-task Triples:} To examine the impact of solely restructuring the input into Open IE Triples but not leveraging graph construction to reduce redundancy, we experiment with a Triples Only baseline. The triples are concatenated to form the input.
    \item \textit{Seq2Seq Multi-task Top 20 Triples:} As an alternative to using graph construction to compress the full set of Open IE Triples, we explore using TF-IDF overlap with the query to find the most relevant information. We select the top 20 triples to concatenate as input. 
\end{itemize}

\begin{table}[t!]
  \centering \small
  \begin{tabular}{ l l c  c  c}\toprule
    \bf{Model} & \bf{Input} & \multicolumn{3}{c}{\bf{ROUGE}} \\
               & \textbf{Length} & \bf{1} & \bf{2} & \bf{L} \\\hline\hline
        Q + D to A*, TF-IDF & avg 850 & \round{28.32} & \round{5.11} & \round{22.75} \\
        Q + D to A, MMR & avg 850 & \round{28.1} & \round{5.0} & \round{22.9} \\
        Multi-task* & avg 850 & \round{28.94} & \round{5.44} & \round{23.12} \\ 
        \hline         
        Multi-task Triples  & 850 & \round{29.04} & \round{5.2}  & \round{23.2} \\ 
        Multi-task Top20 Trip. & avg 570 & \round{28.79} & \round{5.29} & \round{23.22} \\ 
        \hline 
        Q + D to A Graph    & 850 & \round{28.8} & 5.3 & 23.3 \\ 
        Multi-task Graph    & 850 & \round{29.51} & \round{5.6} & \round{23.6} \\ 
        + Top-100 Attention  & 850 & \round{29.70} & \round{5.7} & \round{23.8}\\ 
        + E-MCA & 11K & \round{30.0} & \round{5.8} & \round{24.0} \\ 
        \bottomrule
\end{tabular}
   \caption{\textbf{Answer Generation on ELI5} using Seq2Seq models receiving the \textbf{Q}uestion and a support \textbf{D}ocument (e.g. TF-IDF selection, Triples, Linearized Graph) to produce the \textbf{A}nswer. * denotes results from \cite{fan2019explain}.}
 \label{tbl:full_rouge_answer}
\end{table}

\begin{table}[t!]
  \centering \small
  \begin{tabular}{ l l c }\toprule
    \bf{Model} & \textbf{InputLen} & \bf{ROUGE-L}\\\hline\hline
        T + D to P*  & 500 & 34.2 \\ 
        LM + D-MCA*  & 11K & 36.2 \\ \hline 
        T + D to P & 500  & \round{33.8} \\
        Multi-task & 500  & \round{34.4} \\ 
        Multi-task Graph   & 500  & \round{34.9} \\ 
        + Top-100 Attention  & 500  & \round{35.2} \\ 
        + E-MCA & 11K  & \round{36.5} \\ 
        \midrule 
        LM + D-MCA + MoE-256* & 7.5K & 38.8 \\ 
        \bottomrule
\end{tabular}
   \caption{\textbf{Lead Paragraph Generation on WikiSum CommonCrawl} using Seq2Seq models receiving the \textbf{T}itle and a support \textbf{D}ocument (e.g. TF-IDF ranking, Linearized Graph) to produce the Lead \textbf{P}aragraph.
   * denotes results from \cite{liu2018generating} that use data scraped from unrestricted web search, not the static CommonCrawl version.}
 \label{tbl:wikisum_experiments}
\end{table}

\section{Results} 

We examine the performance of our proposed approach and the choices made in graph construction and modeling. We analyze the quality of the compression created by graph construction and the robustness and interpretability of this process.

\subsection{Linearized Graph Improves Performance}

In Table~\ref{tbl:full_rouge_answer}, we compare our methods to various baselines on the ELI5 dataset. Using MMR to select the most relevant non-redundant input is similar to the TF-IDF baseline from \citet{fan2019explain}. The Seq2Seq Multi-task Triples baseline standardizes the input by forming triples but does not remove redundant triples. It produces marginally better results compared to the baseline Multi-Task model. Sub-selecting to the Top 20 Triples is harmful, as similar text has high TF-IDF overlap with the query so redundant information is selected. Creating the graph structure brings an improvement of around 0.6 ROUGE-1. 

Similar trends are seen for the WikiSum dataset in Table~\ref{tbl:wikisum_experiments}, where graph construction improves the  Multi-task model by 0.5 ROUGE-1. These improvements are statistically significant at the 95\% confidence level. 

For both datasets, a further improvement is seen by using the hierarchical attention mechanism to attend to only the most relevant information in the linearized graph input. For ELI5, it brings an additional 0.2 ROUGE-1 improvement and on WikiSum a 0.3 ROUGE-1 improvement. 

By using MCA to scale Seq2Seq models to encode the entire graph, further gains can be seen. Particularly in information synthesis tasks, prior work has shown the importance of reading more information. \citet{liu2018generating} achieved a 2-point ROUGE improvement by reading 11K tokens instead of 500. In our setting, E-MCA improves our results around 0.3 ROUGE on ELI5 and 1.3 ROUGE on WikiSum. We display random generations from both datasets in the Appendix.

We use human evaluation to compare the Multi-task baseline to the Multi-task Graph + Top-$k$ Attention model. \textbf{57.4\%} of evaluations prefer the Multi-task Graph + Top-$k$ Attention model. We conduct a two-tailed binomial test and find this result is statistically significant with $p = 0.003$.

\begin{table}[t]
    \centering \small
    \begin{tabular}{ l c} \toprule
    \bf{Model} & \bf{ROUGE-1} \\ \midrule
        \multicolumn{2}{l}{\bf{(a) Iterative Removal of Model Components}} \\ \addlinespace
        Multi-task Graph  & \round{29.41} \\
        - Graph Embeddings  & \round{29.12} \\ 
        - KB Completion Multi-tasking  & \round{28.88} \\ 
        - LM Multi-tasking from \cite{fan2019explain} & \round{28.41} \\ \midrule
        \multicolumn{2}{l}{\textbf{(b) Removing Graph Embedding Components}} \\ \addlinespace
        Graph & \\ 
        + Gated Graph Weight + Query Relevance  & \round{28.6} \\
        No Graph Weight Embedding  & \round{28.4} \\ 
        No Query Relevance Embedding  & \round{28.3} \\ 
        No Gating & \round{28.4} \\ \midrule
        \multicolumn{2}{l}{\textbf{(c) Varying Number of Hits in Graph}} \\ \addlinespace
        \multicolumn{2}{l}{Multi-task Graph + Top-$k$ Attention + E-MCA}\\
        with Graph on 5 Search Hits  & \round{28.8} \\
        with Graph on 10 Search Hits  & \round{29.3} \\
        with Graph on 50 Search Hits  & \round{29.6} \\
        with Graph on 100 Search Hits  & \round{29.9} \\ \midrule 
        \multicolumn{2}{l}{\textbf{(d) Varying $k$ in Hierarchical Top-$k$ Atttention}} \\ \addlinespace
        \multicolumn{2}{l}{Multi-task Graph + E-MCA +} \\
        Top-$k=50$ & \round{29.1} \\
        Top-$k=100$  & \round{29.5} \\
        Top-$k=250$  & \round{29.4} \\
        Top-$k=500$  & \round{29.3} \\
        Top-$k=1000$ & \round{29.2} \\ 
        \bottomrule
    \end{tabular}
    \caption{\textbf{Ablations} on the ELI5 Validation Set} 
    \label{tbl:graph_ablation}
\end{table}

\begin{figure*}[t!]
    \centering
    \includegraphics[width=\linewidth]{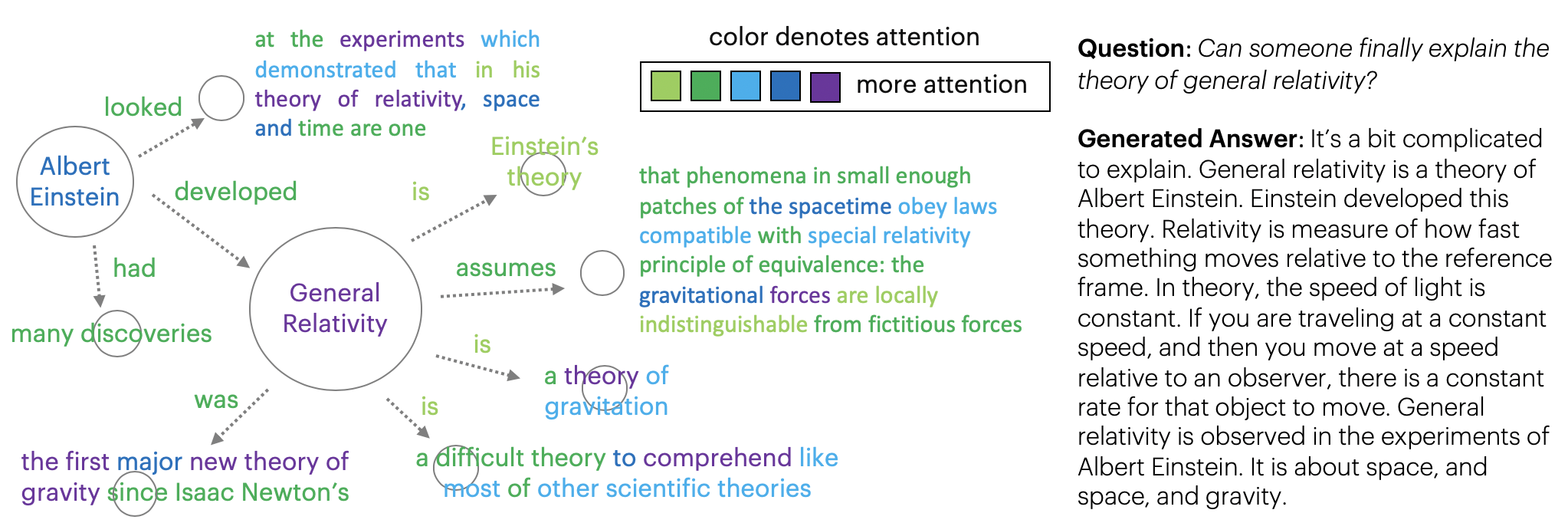}
    \caption{\textbf{Interpretable Attention} of Seq2Seq models on a subgraph when answering a question in ELI5}
    \label{fig:graph_attention}
\end{figure*}

\begin{figure*}[t!]
    \centering
    \includegraphics[width=0.7\linewidth]{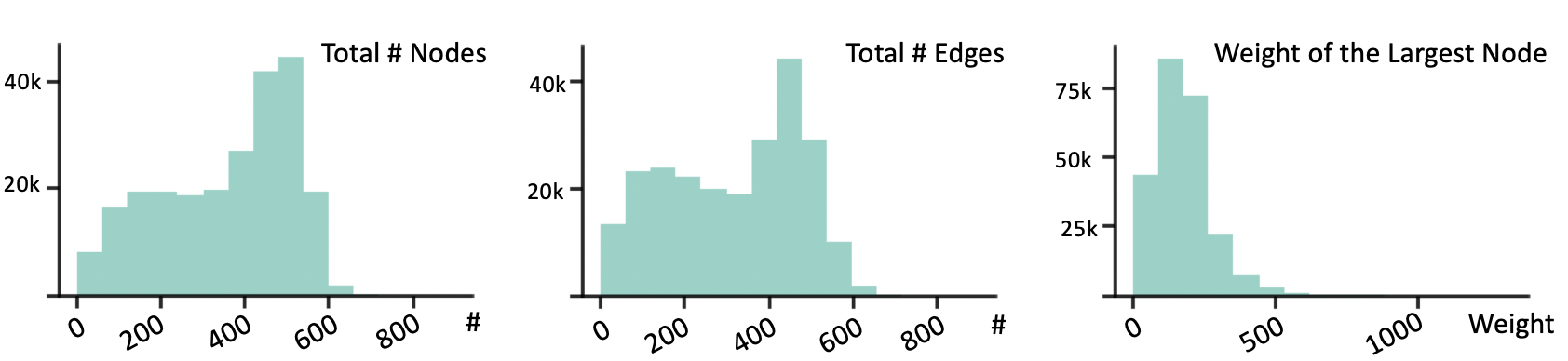}
    \caption{(Left) Distribution of Number of \textbf{Nodes}, (Middle) Number of \textbf{Edges}, (Right) \textbf{Weight} of the Largest Node in graph construction on the ELI5 training set.}
    \label{fig:graph_distribution}
\end{figure*}

\begin{figure}[t!]
    \centering
    \includegraphics[width=\linewidth]{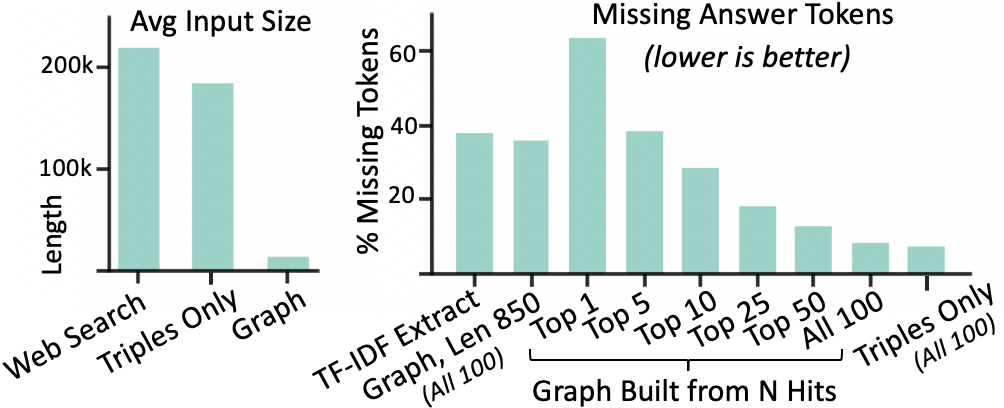}
    \caption{(Left) Graph construction drastically reduces input size by an order of magnitude. (Right) Graph construction encodes more tokens present in the answer compared to TF-IDF extraction and building the graph from more search hits increases answer token coverage. Analysis on ELI5 for both plots.}
    \label{fig:graph_statistics}
\end{figure}

\begin{table}[t!]
  \centering \small
  \begin{tabular}{ l l c }\toprule
    \bf{Model} & \bf{Input} & \bf{ROUGE-1}\\\hline\hline
        Seq2Seq Q + D to A  & TF-IDF  & 28.3 \\
        Seq2Seq Q + D to A  & Web  & 25.9 \\
        + Shuffle           & Web Shuffle  & 24.4 \\ \hline 
        Seq2Seq Q + D to A Graph   & Web &  \round{28.49} \\ 
        + Shuffle           & Web Shuffle  & \round{28.22} \\
        \bottomrule
\end{tabular}
   \caption{Importance of \textbf{Web Search Relevance} on Validation for ELI5, modeling 850 input words.}
 \label{tbl:shuffle_input}
\end{table}

\subsection{Analysis of Modeling Choices} 

\paragraph{Ablation on Model Components} Table~\ref{tbl:graph_ablation}(a) sequentially removes the graph embeddings, the knowledge-base completion multi-tasking, and the multi-tasking from \cite{fan2019explain} and reveals that each of these is important for performance. 

\paragraph{Graph Attribute Embeddings} Table~\ref{tbl:graph_ablation}(b) displays the effect of removing the graph attribute embeddings and gating mechanism. Removing each is slightly harmful, and the combination of all three together provide the best performance. 

\paragraph{More Web Search Documents} Figure~\ref{fig:graph_statistics} (right) shows that graph construction with more web search information is important for answer token coverage. The graph on the top search hit alone is missing 64\% of the answer tokens, but this decreases as more search hits are used.  Table~\ref{tbl:graph_ablation}(c) indicates that this lack of coverage of the answer tokens correlates with generation quality. Models receiving a graph built on the first 5 search hits alone are substantially worse than all 100  hits.

\paragraph{Top-$k$ Attention} Table~\ref{tbl:graph_ablation}(d) shows the effect of the Top-$k$ Hierarchical Attention mechanism for various values of $k$. Attending to too many tokens lowers ROUGE --- for the ELI5 task of writing approximately 130 word answers, attending to 1000 input tokens likely means the model is focusing on irrelevant information and 50 tokens is too few.

\subsection{Graph Improves Answer Token Coverage Despite Compression}

Figure~\ref{fig:graph_distribution} displays the distribution of the number of nodes, edges, and the largest node weight for each local graph built on the ELI5 dataset. The 100 web search results are compressed to a few hundred nodes. By merging redundancy and trimming irrelevant triples from the graph, the input is reduced by an order of magnitude (Figure~\ref{fig:graph_statistics}, left). 

Despite compression, the graph retains more answer tokens than TF-IDF subselection. Figure~\ref{fig:graph_statistics} (right) displays the percentage of answer tokens not present in the input. The TF-IDF Extraction from \cite{fan2019explain} is missing 38\% of tokens. The graph constructed on all 100 web search results is only missing 8.7\% of tokens, but has around 10K words. When analyzing just the first 850 tokens to match the average length of the TF-IDF extraction, the graph is better (only missing 35\% of  tokens). Further, the merging and discarding operations done during graph construction do not have a large effect on answer token coverage --- the full set of triples marginally reduces the percentage of answer tokens missing to 7.3\%  instead of 8.7\%. This indicates that much of the information in the full set triples is redundant and unnecessary for good answer token coverage.

\subsection{Graph Representation is More Robust to Poor Search Relevance Ordering}
We analyze the robustness of our approach to the ordering of web search results in Table~\ref{tbl:shuffle_input}. Instead of constructing the graph from the first web search result to the last, we shuffle the web search results and construct the graph on this shuffled input. We compare this to  modeling the web search results directly (no TF-IDF retrieval) and a model that receives this shuffled web search input. The graph is more robust to shuffling --- as more information can be encoded in the graph due to its compression effect, the search hit ordering is less critical. 

\subsection{Interpretable Attention on Subgraphs}

Figure~\ref{fig:graph_attention} shows an example of the nodes and edges the model focuses upon most when answering a question on ELI5. To construct this visualization, we calculate the top nodes the model attends to and then their top edges. The model attention on a sub-portion of the linearized input can be visualized as an interpretable graph that corresponds well to the model's generated answer. For example, the relationship \textit{General Relativity} $\xrightarrow[\text{is}]{}$ \textit{Einstein's theory} becomes the generated sentence \textit{General Relativity is a theory of Albert Einstein}. 

\section{Conclusion} 

Many open-domain NLP tasks rely upon multi-document input from the web to facilitate tasks such as answering questions or writing summaries, but current approaches struggle to encode the entirely of this information. We propose constructing one knowledge graph per query and show that this method compresses information and reduces redundancy. We show on two abstractive generation tasks that using the linearized graph achieves better performance than TF-IDF retrieval. 

\bibliography{emnlp-ijcnlp-2019}
\bibliographystyle{acl_natbib}

\clearpage
\newpage 
\section{Appendix}

\subsection{Dataset Examples}

We display input and target examples for ELI5 and WikiSum in Figure~\ref{fig:example_eli5} and Figure~\ref{fig:example_wikisum} respectively.

\begin{figure*}[t]
  \small
  \rule{\linewidth}{1pt}
    \noindent \textbf{Question:} Why consumers are still so terrified of genetically modified organisms (GMOs), yet there is little debate in the scientific community of whether they are safe or not. (Scientists are for GMOs)\\  

    \noindent \textbf{Beginning of Web Search: } The controversial safety of gmos and the skepticism of the use of gmos College paper Writing Service URL\_0. diamond chemicals plc the merseyside. appendix f research question for. antisocial personality disorder affects family relations and interactions. The controversial safety of gmos and the skepticism of the use of gmos. Gmo facts what is a gmo genetically modified organisms the safety of gmos is unknown poll: skepticism of genetically modified foods abc news abc news network, 19 june 2015 web fernandez-cornejo, jorge, and seth james wechsler. The controversy over gmos, particularly in food continues: scientists are split pros and cons of gmo's september 28 and environmentalists and consumer groups remain unconvinced about the safety of gmos. The controversy around genetically modified foods on the surface, food evolution is a resetting of the controversial conversation around genetically modified organisms (gmos) we just ask people by a show of hands to tell us are they concerned about gmos for their own safety or the. When gmos are the movie star can documentaries on controversial science be entertaining and the message is that gmo food is safe to eat and that naysayers are. Genetically modified organisms (gmos) the top five anti-gmo tropes gmos are genetically modified organisms the evidence on gmo safety by ramez naam. Genetically modified organisms what are gmos with the use of gm technology, pure and safe equivalents can be produced using gmos and industrial scale. Here's a bullet-point summation of what nathanael johnson learned about gmos in 2013 20 gmo questions: animal, vegetable, controversy by pretty darn safe. The controversy surrounding genetically modified organisms what do we tolerate as far as detrimental this would be a profound service to scientific skepticism with regards to gmos current gmos are safe to eat- however [...]\\  
    
    \textbf{Target Answer:} There is little difference in essence between what is called GMO now and the techniques we have been using to domesticate and cultivate the food in the past. Its an arbitrary line that's been drawn in the sand and people fail to realize this often. That being said I think it is more then wrong the patenting of crops and again even more then wrong to genetically modify crops to not have viable seeds so that seed-washing can't be used to grow the next crop. So the real god damned issues are being masked and lost by this retarded polemic between GMO and more conventional genetic modification of organisms. \\

  \rule{\linewidth}{1pt}
  \caption{Example of ELI5 Dataset Input-Output}
  \label{fig:example_eli5}
\end{figure*} 

\begin{figure*}[t]
  \small
  \rule{\linewidth}{1pt}
    \noindent \textbf{Title:} dwight h. perkins (economist)\\  

    \noindent \textbf{Beginning of Web Search: } memorial minute adopted by the faculty of arts and sciences, harvard university: abram bergson, john kenneth galbraith, carl kaysen, raymond vernon, dwight h. perkins, chairman. ed was a generous and understanding man who could see the good qualities in people and bring them out. he was always determined and persistent, which meant that he also saw to completion what he started . the list of projects, large and small, that he led is long. in 1946 he was one of the authors of the speech of secretary of state james byrnes in which the secretary announced the return of responsibility for the german economy to the germans. in 1956, he, together with ray vernon, did a pioneering study of the new york metropolitan region, a study that tried to identify the economic, political and social forces that were shaping that vast urban area. at the time the problems of urban areas were mainly the concern of architects and philosophers; almost no economist and only a few political scientists knew much or cared greatly about such issues. the studies that resulted provided rich fare for a generation of urban planners to follow. mason, edward s.(1899-1992) | harvard square library  home » biographies » mason, edward s.(1899-1992) mason, edward s.(1899-1992) edward sagendorph mason was a member of the first parish unitarian church in harvard square who exemplified liberal religion in both thought and action. in addition to his notable contributions to the science of government, he served as chairman of the sloan commission on cable communication which issued recommendations for the future, on the cable.  [...]\\  
    
    \textbf{Target Lead Paragraph:} dwight heald perkins (born in chicago, illinois in 1934) is an american academic, economist, sinologist and professor at harvard university. he is the son of lawrence bradford perkins, architect, and margery blair perkins and the grandson of dwight heald perkins, the architect, for whom he was named. he married julie rate perkins in 1957 and they have three adult children. == early life == perkins earned an undergraduate degree at cornell university in 1956. after two years military service in the us navy, perkins resumed his studies at harvard. he earned a ma in economics in 1961 and a ph.d. in economics in 1964. his doctoral thesis was ``price formation in communist china''. == academic career == perkins' teaching career at harvard began when he was still a graduate student, and continued uninterrupted through 2006 when he became a research and emeritus professor. he was a member of both the department of economics of the faculty of arts and sciences at harvard and of the harvard kennedy school. [...] \\

  \rule{\linewidth}{1pt}
  \caption{Example of WikiSum Commoncrawl Dataset Input-Output}
  \label{fig:example_wikisum}
\end{figure*}

\subsection{Generation Examples}

We display examples of model generations for ELI5 and Wikisum, selected randomly, in Figure~\ref{fig:examples_eli5} and Figure~\ref{fig:examples_wikisum_generation}. Similar to \cite{liu2018generating}, we observe that our models are also capable of generating full Wikipedia pages. We display randomly sampled examples of  Wikipedia article generation in Figure~\ref{fig:examples_wikipedia_generation}.

\begin{figure*}[t]
  \small
  \rule{\linewidth}{1pt}
    \noindent \textbf{Question: Why is touching microfiber towels such an uncomfortable feeling?}\\  
    
    \textbf{True Answer:} Do you mean the kind of cloths used to clean glasses and lenses? I've never noticed any uncomfortable feeling myself, but I do find touching certain cleaning cloths can be quite uncomfortable. There's a brand called ``e - cloth'' which market themselves as not needing any cleaning supplies. Basically the threads have tiny little hooks and barbs in them, so dragging them over skin makes you feel like it 's catching against you.\\
    
    \textbf{Generation:} The microfiber is made up of a bunch of tiny fibers that are attached to them. When you touch them, the fibers that make up the microfiber are attracted to each other. When they touch, they're attracted to the other end of the fiber, which is what makes them feel uncomfortable.  \\  

    \vspace{2mm} \textbf{Question: What happened to needing ``www'' in front of an address?} \\ 
    
    \textbf{True Answer:} Explaining like If you were 5: You know when you send a package in the mail, you write the address and maybe, If the package is ``Fragile'' you'll stamp ``FRAGILE'' all over it. Why don't you also stamp ``NOT FRAGILE'' on the packages that are not fragile? Because It's implicit. If you don't say anything the mailman knows It's not fragile. Same thing goes for the WWW. Since you can have multiple services on your domain (ex: URL\_0 which means you want to share files; or URL\_2 which means you want to access the mail server on that domain) in the beginning you'd also write ``URL\_1'' to state that you wanted to access the HTML of the server. As the web evolved and 99\% of the average user wants to access the HTML version of the website, website owners no longer require you to type ``URL\_1'' instead, If you type ``URL\_3'' they assume you want to access the HTML version. Just like you don't need to stamp ``NON FRAGILE'' on your package, you don't need to type ``WWW''. Everyone assumes that's what you want If you say nothing otherwise.\\
    
    \textbf{Generation:} The ``www'' is a subdomain name that is used to refer to a web page. It's a name used to describe the web page that you're looking for. For example, if you want to go to URL\_0, you can go to URL\_1, and then go to URL\_2. The www is the name of the site that you are looking for, and it's the name that you were looking for in the first place. So if you wanted to go back to URL\_2, you could go to URL\_4, and go to URL\_3, and the www would be the same as the www, but it would be a different name for the site. So, if the www was the same, you would be looking for a different site, and you wouldn't be looking at the same site, so you would have to go through the www. If the www were the same and you weren't looking for the same page, you'd have to look for the different sites, and so on. But if you were to go from URL\_1, you might have to use the www as a reference to the site, but you'd still have to have a www.\\ 
    \\
    
    \vspace{2mm} \textbf{Question: Why do we have nightmares when stressed out?}\\
    
    \textbf{True Answer:} I can tell you that your mind, and in particular your dreams, are subject to the input that it the journey of your life, as you have interpreted it. Yours dreams tend to be your subconscious representing issues you have within, whether it be past or present, with things and people that are sort of ``preinterpreted symbology'' from your deep and ever power subconsciousness. All your nightmares are just dreams, and your inner mind knows exactly can mess with you the most, and sometimes it brings that to light, it usually being an issue you may be stressed about.\\
    
    \textbf{Generation:} When you're stressed, your brain releases a hormone called cortisol. This hormone is released by your adrenal glands, and is released when you are stressed. This is why it's called ``fight-or-flight'' mode. When your body is stressed, it releases hormones that make you feel more stressed. The reason why you don't feel stressed is because your brain is trying to get rid of the stress hormone, which causes you to feel stressed, and you're less likely to get stressed. If you're really stressed, you're probably going to be more stressed than if you weren't stressed.\\
  \rule{\linewidth}{1pt}
  \caption{Random Examples of Question-Generated Answer Pairs in ELI5}
  \label{fig:examples_eli5}
\end{figure*} 

\begin{figure*}[t]
  \small
  \rule{\linewidth}{1pt}
    \noindent \textbf{Title: wylie rogers} \\
    
    \textbf{True Paragraph:} wylie rogers (born march 16, 1985) is an american ice hockey goaltender. he played with the rio grande valley killer bees of the central hockey league. prior to turning professional, rogers attended the university of alaska fairbanks where he played four seasons of ncaa men's division i ice hockey with the alaska nanooks men's ice hockey team. rogers played the 2010-11 season in the netherlands with the heerenveen flyers.\\

    \textbf{Generation:} wylie rogers is an american football coach and former player. he is currently the head football coach at the university of north carolina at chapel hill. rogers played college football at north carolina state university. he was a member of the north carolina tar heels men's football team that won the ncaa division i football championship in 2012. he also played for the university's men's basketball team. after his playing career ended, rogers was hired as the head coach at unc chapel hill in 2013. he led the tar heels to their first ncaa division ii championship in 2015. \\ 
    
    \vspace{2mm} \textbf{Title: mustabad}  \\
    
    \textbf{True Paragraph:} mustabad is a village and mandal in siricilla rajanna district in telangana state of south-east india. musthabad is the main town for the mustabad mandal. musthabad is located 26,160; km distance from its district main city, siricilla. it is located 128,160; km distance from its state main city, hyderabad. == geography == mustabad is located at 18.2787 \degree n 78.7108 \degree e.\\
    
    \textbf{Generation:} mustabad is a village in the indian state of telangana. it is located in the region of mustabad mandal.  mustabad was founded in the early 1970s by a group of local businessmen, who wanted to create a village for the people of the region. the village is named after the mustabad river, which is a tributary of the larger river.  according to the 2011 census of india, mustabad had a population of 1,050. \\ 
    
    \vspace{2mm} \textbf{Title: wolf valley} \\
    
    \textbf{True Paragraph:} wolf valley is a graben rift valley in the elsinore trough, in western riverside county, california. wolf valley is the southernmost of the graben valleys making up the elsinore trough, created by the elsinore fault zone. it lies between the wildomar fault on the east, at the foot of the temecula basin and the willard fault on the west, at the foot of the santa ana mountains. wolf valley lies south of temecula creek. the southern end of the graben valley is where the elsinore fault zone changes the direction of strike along the southern side of agua tibia mountain , northeast of pala mountain. the valley is drained by pechanga creek and its tributaries, itself a tributary of the santa margarita river. the pechanga indian reservation and part of the city of temecula is located in this valley. head of wolf valley 33 \degree 26 ′ 38 ″ n 117 \degree 03 ′ 46 ″ w mouth of wolf valley 33 \degree 28 ′ 26 ″ n 117 \degree 07 ′ 03 ″ w\\
    
    \textbf{Generation:} wolf valley is a valley in the u.s. state of california. it is located in the foothills of the sierra nevada. the valley was named after the wolf river, which flows through the valley. wolf valley was the site of the battle of wolf valley, which took place in the late 19th and early 20th centuries. the battle was fought in the early 20th century, and the valley now holds the wolf valley war memorial.\\
  \rule{\linewidth}{1pt}
  \caption{Random Examples of Lead Paragraph Generation in WikiSum}
  \label{fig:examples_wikisum_generation}
\end{figure*} 

\begin{figure*}[t]
  \small
  \rule{\linewidth}{1pt}
    \noindent \textbf{kokal}  kokal is a village in kollam district in the state of tamil nadu, india.  \\
    == demographics == \\ as of 2001 india census, kokal had a population of 1,051 \\
    == transport == \\ kokal railway station is located at kokal junction. the station is well connected to other parts of the state by road. the nearest airport is at tiruchirappalli.\\
    == education == \\ the primary and secondary school education is imparted by government, aided and private schools. the medium of instruction followed by different schools are english, telugu, malayalam and other languages. \\
    == in popular culture == \\ in the film, the character of the protagonist of the film is played by a character named kokal. in the movie, the protagonist is portrayed by a. k. ramachandran, who is also the protagonist. \\
    == in other media == \\ the village has a tv station called kokal - kokal which is a part of the local tv station. the village also has a radio station called kollam radio which is the local radio station of the same name. the radio station is also called ``kollam fm'' and is known as the station' s main radio station. \\
    
    \noindent \textbf{donald milne} donald milne is a canadian politician. he is a member of the legislative assembly of british columbia, representing the electoral district of vancouver. milne was born in vancouver, british columbia. he was elected to the british columbia house of commons in the 2014 provincial election, defeating liberal candidate and former mayor of vancouver, david campbell. \\ 
    == electoral record == \\ by - election results == == 2014 == == 2015 == == 2016 == == 2017 == == 2018 == == 2019 == == 2020 == == future == milne will be a candidate for vancouver city council in the 2018 provincial election. \\
    == personal life == \\ milne lives in vancouver with his wife and two children. \\
    == awards and honours == \\ milne was a recipient of the queen elizabeth ii diamond jubilee medal from the queen's birthday honours in 2014. he also received the order of the british empire in the 2016 birthday honours for services to british columbia and the community in british columbia in the 2015 birthday honours. milne is also a fellow of the royal canadian academy of arts and sciences and the royal society of canada (rbc) and is a past president of the vancouver chapter of the rbc society of arts and sciences. in 2016, milne was awarded an honorary doctor of laws degree by the university of vancouver in recognition of his service to the community and his contribution to the development of vancouver's arts and science as a leading figure in the vancouver art scene. in 2017, he was awarded a lifetime achievement award by the vancouver arts council for his work on vancouver art and science, as well as an honorary doctorate of laws (honoris causa) from the royal college of art and sciences, university of toronto in the 2017-18 academic year. \\
    
    \noindent \textbf{decimiana} decimiana is a genus of flowering plants in the legume family, fabaceae. species include: decimiana acuminata (f. muell.) benth. ex benth - ecuador, peru, bolivia, brazil decimiana angustifolia benth. \\
    == species == \\ there are currently two species in this genus, decimiana acutifolia and decimiana costata, which are native to ecuador and peru, but are now considered a separate species. the following is a list of species of this genus in the world, including:  decimiana borealis benth (ecuador, colombia), decimiana cunninghamii benth (ecuador, bolivia) some of these species are listed in the iucn red list of threatened species. the others are listed as endangered species by the international union for conservation of nature (iucn), and are listed on the endangered species list by the united states department of agriculture (usda). a few of the species of a. albiflora are listed by the iucn as endangered or threatened species, but they are now listed as critically endangered. \\ 
    == the others == \\ the other members of the family are listed under the iucn's red list, but the other two are listed separately as endangered. the other three are listed together as endangered and listed as vulnerable species. \\
  \rule{\linewidth}{1pt}
  \caption{Random Examples of Full Wikipedia Generation in WikiSum.}
  \label{fig:examples_wikipedia_generation}
\end{figure*} 

\subsection{Implementation Details} 

\paragraph{Training} We train models using the Transformer Big architecture with the following parameters: dropout $0.1$, attention dropout $0.1$, an inverse square root learning rate schedule with initial learning rate $1e^{-7}$ for warmup, $4000$ warmup updates, and a minimum learning rate of $1e^{-9}$. We train with fp16 and small batchsize to scale to longer sequences. To offset the batchsize decrease, we increase the update frequency of the gradient updates. 

\paragraph{Generation} We generate using beam search of size $5$. We tune a minimum and maximum length at validation time, and use minimum length $200$ and maximum length $500$. We use a n-gram constraint of $n=3$ following \cite{fan2019explain}.  

\end{document}